\documentclass{article} 

\usepackage[margin=1in]{geometry}

\usepackage{amsmath,amsfonts,bm}

\def\eqref#1{equation~\ref{#1}}

\def\1{\bm{1}}

\DeclareMathAlphabet{\mathsfit}{\encodingdefault}{\sfdefault}{m}{sl}
\SetMathAlphabet{\mathsfit}{bold}{\encodingdefault}{\sfdefault}{bx}{n}

\usepackage[T1]{fontenc}
\usepackage{graphicx}
\usepackage{wrapfig}
\usepackage{listings}
\usepackage{hyperref}
\usepackage{url}
\usepackage{minted}
\usepackage{float}
\usepackage{longtable}

\usepackage{natbib}
\title{AutoHarness: improving LLM agents by automatically synthesizing a code harness}

\author{Xinghua Lou, Miguel L\'azaro-Gredilla, Antoine Dedieu, \\Carter Wendelken, Wolfgang Lehrach, Kevin P. Murphy\\
Google DeepMind\\
\texttt{\{xinghua,lazarogredilla,adedieu,cwendelken,wpl,kpmurphy\}@deepmind.com}
}

\newcommand{\eat}[1]{} 

\begin{document}

\maketitle

\begin{abstract}
Despite significant strides in language models in the last few years,
when used as agents, such models often try to perform actions that are not just suboptimal for a given state, but are strictly prohibited by the external environment.
For example,
in the recent Kaggle GameArena chess competition, 78\% of Gemini-2.5-Flash losses were attributed to illegal moves. 
Often people manually write "harnesses" around LLMs
to prevent such failures.
In this paper, 
we demonstrate that Gemini-2.5-Flash can automatically synthesize such a code harness,
using a small number of rounds of
iterative code refinement given feedback from the (game) environment.
The resulting harness
 prevents  all illegal moves in 
 145 different TextArena games (both 1-player and 2-player), enabling the smaller Gemini-2.5-Flash model to outperform larger models, such as Gemini-2.5-Pro.
 Pushing our technique to the limit, we can get
 Gemini-2.5-Flash to generate the entire policy in code,
 thus eliminating the need to use the LLM at decision making time. The resulting code-policy
 receives a higher average reward than
 Gemini-2.5-Pro and GPT-5.2-High
 on 16 TextArena 1-player games.
 Our results
 show that  using a smaller model to synthesize a custom code harness (or entire policy) can outperform a much larger model,  while also being more cost effective.
\end{abstract}

\eat{
\begin{abstract}
Despite significant strides in language models in the last few years regarding reasoning and planning,  the performance of current flagship models in board games can be surprisingly disappointing. 
In the recent Kaggle GameArena chess competition, 78\% of Gemini-2.5-Flash losses were attributed to illegal moves. While a hand-crafted harness blocking illegal moves would offer a performance boost, it lacks generality. In this work, we demonstrate that Gemini-2.5-Flash can evolve a programmatic harness for TextArena games, effectively preventing illegal moves. With this approach, we outperform a superior model, Gemini-2.5-Pro, in both 1-player and 2-player games. Additionally, when applying this method to evolve a programmatic policy using Gemini-2.5-Flash, we achieve a high average reward than GPT-5.2-High on TextArena 1-player games. Our results show that letting a ``good-enough'' model write its own custom harness (or even policy) can outperform a better model on specific tasks while being more cost effective.
\end{abstract}
}

\section{Introduction}

Large language models (LLMs) have demonstrated remarkable capabilities in code synthesis
and solving math problems
(see e.g., \cite{chervonyi2025gold, huang2025winning}).
However, their planning and reasoning performance
can be brittle
(see e.g., \citep{Valmeekam2023,petrov2025proof}.
For example,
in the recent Kaggle GameArena \citep{kaggle_game_arena_2025} chess competition, 78\% of losses by Gemini 2.5 Flash were attributed not to strategic blunders, but to simple illegal moves.

This failure mode highlights a disconnect between the model's apparent understanding of the game and its 
ability to actually follow the rules
(see e.g. Fig. A16 in
\citep{Ruoss2024}).\footnote{
The general problem of knowing which actions
are valid in a given state is called
the "action applicability" problem,
and has been studied in the AI planning
community
\citep{Kokel2025}.
}
Traditional approaches to mitigate this involve fine-tuning on game trajectories or using hand-coded harnesses that verify the validity of a move. Fine-tuning LLMs, particularly at the scale of current flagship models, is neither fast nor cost effective, and can degrade model performance on other tasks, e.g.\ instruction following. Hand-designed harnesses are brittle and labor-intensive, requiring additional work for every new game. A more scalable solution --- which we pursue in this paper --- is to leverage the LLM’s own code-generation capabilities to bridge this gap.

An agent is often defined as the combination of a specific LLM and a harness that acts as the ``glue'' or ``plumbing'' between the model and the task that needs to be solved. In this work, we propose ``code as harness'', a framework where the LLM itself completes the agent by coding its own harness. In its simplest incarnation, the harness can be seen as a control loop that calls the LLM and rejects unacceptable answers. The definition of what is acceptable is itself learned. This essentially results in a rejection sampler for LLMs in which the conditioning is learned based on the task.

We formulate the generation of this harness as a search problem over the space of programs. Unlike simple iterative prompting, we employ a tree search guided by Thompson sampling \citep{tang2024code} to efficiently explore the landscape of potential harnesses. In this setup, the LLM acts as a mutation operator, proposing refinements to the code based on feedback from execution. The search algorithm balances exploration (trying distinct logic structures) and exploitation (refining a partially working harness) to converge on a robust control loop. The harness template can be more constrained (e.g., a fixed rejection sampling loop where we only learn a conditioning function with signature \texttt{def is\_legal\_action()}), or less so, with maximum flexibility resulting in a code-as-policy setup \citep{liang2023code} in which code proposes the next action directly and no LLM calls are needed at execution time.

\section{Related work}
\paragraph{LLMs for game playing and reasoning} The use of LLMs as agents in game environments has been widely studied, ranging from text-based adventure games to complex strategy games like Minecraft and chess \citep{shinn2023reflexion, wang2023voyager}. Early works focused on ``chain-of-thought'' prompting \citep{wei2022chain} to improve strategic planning. However, recent benchmarks reveal that even advanced models struggle with state tracking and validity in strictly defined environments \citep{valmeekam2023planning}. Techniques like ``tree of thoughts'' \citep{yao2023tree} utilize search during inference to simulate lookahead, but they rely on the LLM's internal world model, which is prone to hallucination regarding valid transitions. Our work differs by offloading the state-transition 
validity checker
to an external, verifiable program rather than relying on the model’s internal simulation.
LLMs can also be used to generate 
code for the entire
state transition function (i.e. world model) for
a game \citep{lehrach2025code}, but that is unnecessarily onerous for complex games in which a comparatively simple strategy can be applied.
In addition, this approach does not
leverage
the strategic abilities of the LLM to select between valid actions.

\paragraph{Code as policy} Our approach builds upon the growing body of work using code generation for action planning. Voyager \citep{wang2023voyager} demonstrated that LLMs could continuously learn Minecraft skills by storing executable code in a library. Similarly, Eureka \citep{ma2024eureka} showed that LLMs could perform evolutionary search to generate reward functions for reinforcement learning. Closer to our work, code as policies \citep{liang2023code} formulated robot control directly as code generation.
Our approach is related,
but uses 
\emph{iterative code refinement},
based on tree search and rich environment feedback,
to generate a hybrid code+LLM harness.

\paragraph{Refinement and search} As mentioned, iterative refinement is crucial for code generation. Reflexion \citep{shinn2023reflexion} introduced a verbal reinforcement learning loop where agents reflect on failure logs. In the domain of program synthesis, methods like AlphaCode \citep{li2022competition} utilize large-scale sampling and filtering, whereas AlphaEvolve \citep{novikov2025alphaevolve} applies an evolutionary algorithm to entire codebases using an LLM as a mutation function. Our method integrates these concepts into a structured tree search using Thompson sampling, following \citep{tang2024code}, but applies it in an online, multi-turn setup, 
where the goal is to create a code harness.

\section{Method}

\begin{figure}
    \centering
    \includegraphics[width=0.666\textwidth]{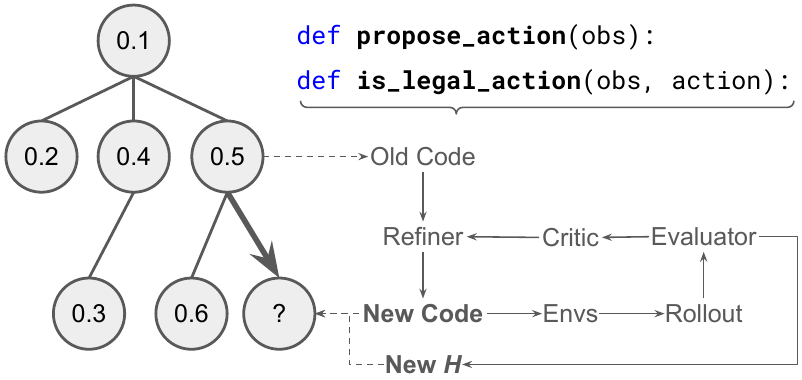}
    \caption{Code-as-harness learning process.}
    \label{fig:approach} 
\end{figure}

Inspired by  \cite{tang2024code},
our approach maintains multiple code hypotheses in a tree structure, and uses Thompson sampling
to choose which node to refine next,
where the heuristic value for each node  is the average legal move accuracy.
The refinement (gradient-free code optimizer)
is done with a base LLM, given feedback from the environment (critic) about whether the previous attempted moves were legal or not, and what reward they produced (if any), see Fig.\ref{fig:approach}. If \texttt{is\_legal\_action()} returns \texttt{True} but the action is invalid, we refine both functions; while if \texttt{is\_legal\_action()} returns \texttt{False} and the action is invalid, we only refine \texttt{propose\_action()}.

\eat{
We cast the code-as-harness learning problem as tree search over the code space, with online feedback. Our approach resembles online, 
on-policy
reinforcement learning without ``gradients'' (Fig.~\ref{fig:approach}) -- the code is the target ``policy'', a critic provides feedback as the ``gradient'', and an LLM refiner acts as the ``optimizer''. Inspired by \cite{tang2024code}, the selection of the base node is based on Thompson sampling, where the heuristic value $H$ is the average legal move accuracy.
}

We can use this approach to generate  different kinds of code harnesses:
 \textbf{harness-as-action-filter} calls \texttt{propose\_action()} to generate a \emph{set} of legal moves, and leverages the LLM to  rank them (potentially using chain of thought reasoning);
 \textbf{harness-as-action-verifier} first calls the LLM to generate an action, verifies it by \texttt{is\_legal\_action()}, and, if invalid, repeats the process with a new prompt that includes an ``illegal action'' warning message;
  \textbf{harness-as-policy} uses code to choose the action;
  the code could in principle call an LLM,
  but in our setting,
  the policy just uses primitive Python functions and standard libraries such as numpy,
  so we do not need to invoke an LLM at inference time.
In this paper, we mostly focus on the harness-as-action-verifier,
but in Sec.\ref{sec:harness-as-policy},
we also report preliminary
results on harness-as-policy.


\section{Experimental results}

For our experiments we select
all the 1-player (1P) and 2-player (2P) games
from TextArena \citep{guertler2025textarena}, a large collection of complex and diverse text games,
but exclude the 9 games
whose action space is free-form text / dialog
(such as "Mafia" and "Codenames").
This leaves us with 145 games, including well-known games
--- such as  Chess, Checkers, Blackjack and Sudoku --- as well as novel variants of these games.
A full list of the games we use is in 
Appendix Table~\ref{tab:long}.

\eat{
We select long-horizon games --- such as 
 Chess, Checkers, Blackjack and Sudoku --- but exclude dialog games,
 such as Mafia and Codenames,
 which allow arbitrary text as actions.\footnote{
 This means we cannot apply our method
 to 3 of the 4 TextArena games used in the
 \url{https://www.mindgamesarena.com/} competition held at NeurIPS 2025.
 }
 See the appendix for details of our chosen games.
}

To make the problem more challenging for our harness,
we modified some games by manually
removing any form of ``Available Moves'' hints in the observation string
(see Appendix Sec.~\ref{sec:remove-legal-chess} for an example).
We believe this better reflects many real-world scenarios where the agent needs to deduce legal actions from environmental feedback,
rather than being told them explicitly.
(Without this modification, the harness can just copy the list of legal actions from the prompt. This gives better results, but we show that it is unnecessary.)

\subsection{Training}

\begin{figure}[h]
    \centering
    \includegraphics[width=0.666\textwidth]{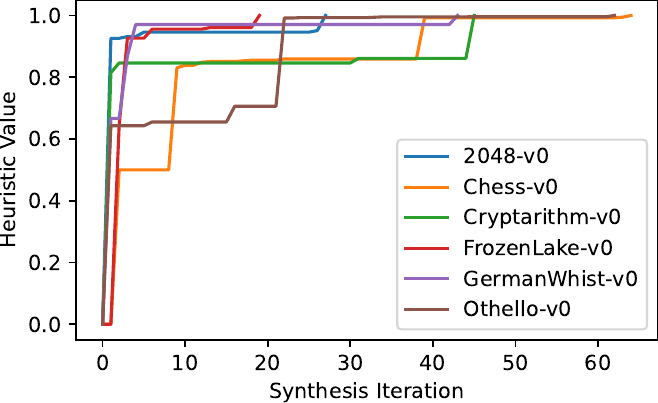}
    \caption{Fraction of legal moves vs number of code refinements for a selection of 6 games.}
    \label{fig:training}
\end{figure}

Our training setup
(for harness-as-action-verifier) 
is as follows. At each iteration, we use 10 parallel environments and roll out to at most 1000 steps (with auto environment resetting). Rollout is terminated whenever an illegal move is made by the code or code execution fails. At most 5 failed steps are sampled and fed to the \textit{Critic}, which consolidates various types of errors. These steps with error messages, together with the original code, are fed into the \textit{Refiner} to generate new (hopefully improved) code. We set heuristic weight to 1.0 for Thompson sampling. Training ends when the heuristic value (i.e. the legal action success rate) reaches 1.0, or we time out.
We use Gemini-2.5-Flash for training.

On average, training ends after 14.5 tree search iterations, while 19/32 games end in less than 10 iterations. The games
that required the most number of LLM calls to learn are  are GermanWhist-v0 (2P), Cryptarithm-v0 (1P), Othello-v0 (2P) and Chess-v0 (2P),
as shown in Fig~\ref{fig:training}.
We measure the accuracy of the action filter by applying it
to novel test rollouts (of length 1000, across 10 random random seeds per game), and measuring the fraction of legal actions.
We achieved 100\% legal action success rate for all the games
as shown in Appendix Table~\ref{tab:long}.
See Appendix Sec.~\ref{sec:snippets} for examples of the generated code harness.


\subsection{Evaluation}

We now turn to evaluating performance of agents during actual game play.
For reasons of efficiency, we focus our results on 16 1P games and 16 2P games, rather than using all 145 games.
We evaluate the following agents:
Gemini-2.5-Flash, Gemini-2.5-Pro and Gemini-2.5-Flash+Harness (ours).\footnote{
Note that our method first uses
an LLM 
(here Gemini-2.5-Flash) to generate the action verifier code harness, and then uses this harness to filter proposals from 
the same LLM.
}
We use the same optimized prompt in all experiments. 
For 1P games, we run 20 matches and use the reward as the evaluation metric. 
For 2P games, we run 40 matches with random seeds, split evenly between our method being the first or second player, and we use the average win/draw/loss rate as the evaluation metrics. 

\begin{figure}[h!] 
    \centering 
    \includegraphics[width=0.999\textwidth]{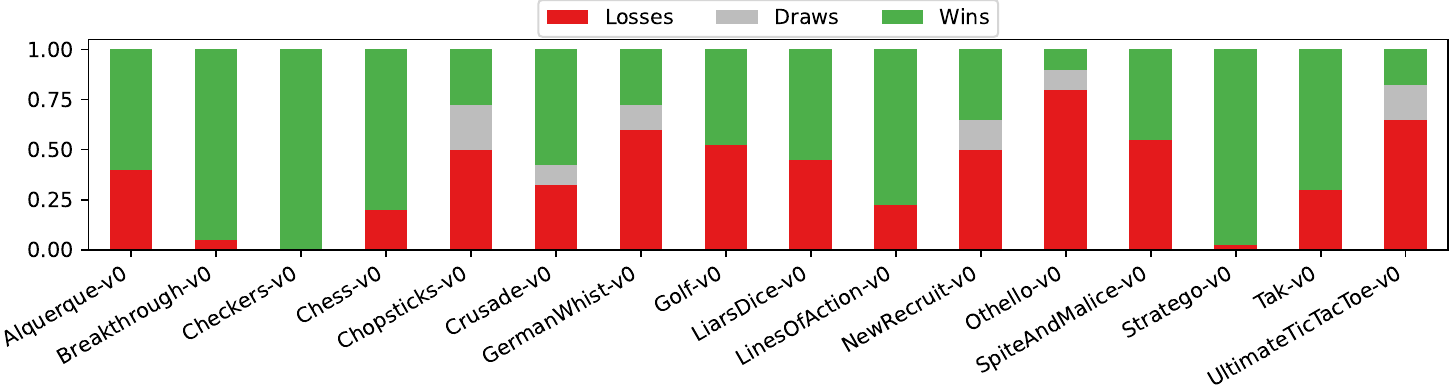}
    \vspace*{-8mm}
    \caption{Win/lose/draw rate of our method vs Gemini-2.5-Pro for each of the 16 2P games.}
    \label{fig:2p-results} 
\end{figure}

We show results for 2P games in Fig.~\ref{fig:2p-results}.
We see that our approach enables a much smaller Gemini-2.5-Flash to win 9/16 games (overall win rate of 56.3\%)
against a much larger Gemini-2.5-Pro
(overall win rate of 38.2\%).
When playing against (vanilla) Gemini-2.5-Flash, we win 12/16 games, and the overall win rate rises to 64.8\%.

\begin{figure}[h!] 
    \centering 
\includegraphics[width=0.999\textwidth]{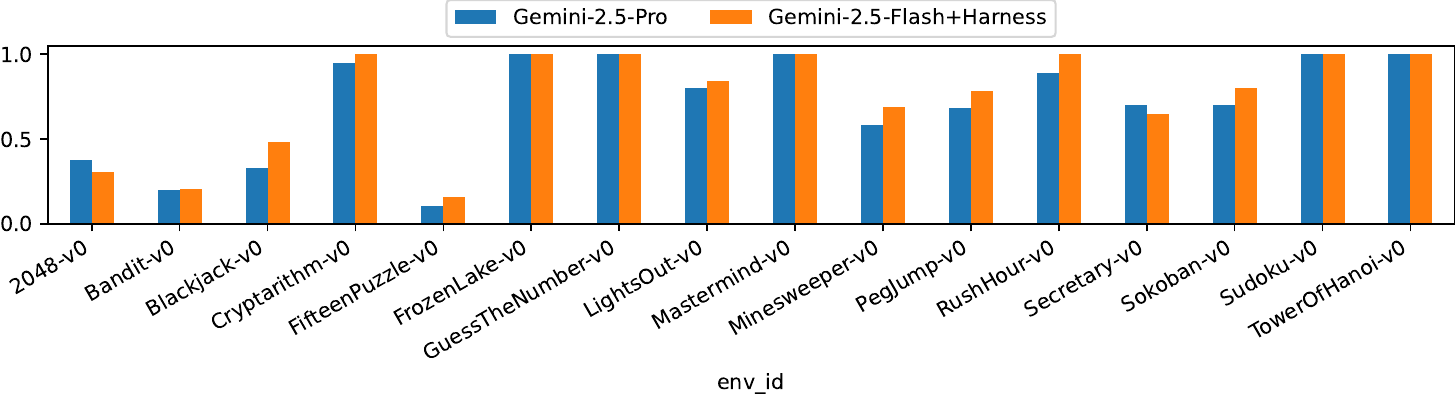} 
    \vspace*{-8mm}
    \caption{Average reward of our method and Gemini-2.5-Pro for each of the 16 1P games.}
    \label{fig:1p-results} 
\end{figure}

We show results for 1P games in Fig.~\ref{fig:1p-results}. 
We see that our approach achieves a higher reward than Gemini-2.5-Pro in 8/16 games, and ties in 5/16 games. On average, we achieve 0.745 reward, in comparison to 0.707 (Gemini-2.5-Pro) and 0.673 (Gemini-2.5-Flash).


\subsection{Harness-as-Policy}
\label{sec:harness-as-policy}

As an extreme case, we consider learning the entire policy as code, dispensing with the need to use an LLM at test time.
We evaluate this on 16 1P games (since it is much harder to learn an entire policy in code form for 2P games\footnote{
Two-player games require strategic reasoning about the opponent's policy
which often requires MCTS-like methods at run time
(see e.g., \citep{Duan2024}).
While in principle our code synthesis method could generate such a policy,
it would also need to learn a code world model to search over, as in 
\citep{lehrach2025code},
which is challenging for text games.
}.) 
In addition to the above agents, 
we evaluate three new agents: GPT-5.2 (no thinking), GPT-5.2-High (high thinking) and Harness-as-Policy (ours).
All agents are evaluated 20 times per game, as before,
except for  GPT-5.2 and GPT-5.2-High,
which are repeated 10 and 5 times, for cost reasons.

\begin{figure}[h!]
    \centering
    \includegraphics[width=0.666\textwidth]{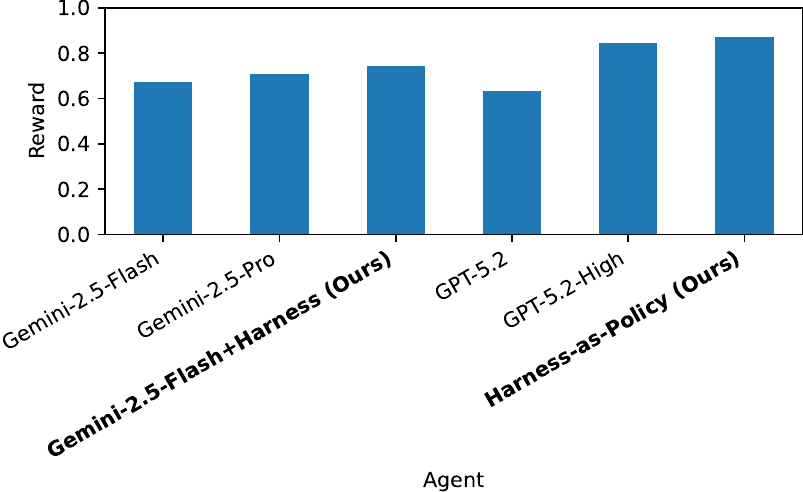}
    \caption{Average reward of different agents across 16 TextArena 1P games.}
    \label{fig:as-policy} 
\end{figure}

For training, we modify the heuristic value to include the reward.
Specifially we set $H=0$ if an illegal action is taken,
and $H=0.5 + 0.5r$ otherwise,
where $r \in [0.0, 1.0]$ is the environment reward,
which is only available at the end of the trajectory (sparse reward setting).
\eat{
\begin{align*}
H =
\begin{cases}
    0.0 & \text{if illegal action} \\
    0.5 + 0.5r & \text{otherwise}
\end{cases}
\end{align*}
, where the reward $r \in [0.0, 1.0]$ is only available at the end of the trajectory.
}
We train Harness-as-Policy using our code synthesis method with Gemini-2.5-Flash to a maximum of 256 iterations. On average, training takes 89.4 iterations and achieves a heuristic value of 0.939.

As shown in Fig.~\ref{fig:as-policy}, our approach achieves the highest average reward (0.870), outperforming all other agents including 
GPT-5.2 (0.635),
Gemini-2.5-Pro (0.707),
and
GPT-5.2-High (0.844).
Per game, we win 3/16 games while GPT-5.2-High wins 5/16, and we tie the remaining 8/16 (details in the appendix).
Since Harness-as-Policy generates pure (Python) code, our  test time cost is nearly zero, while the GPT-5.2 and GPT-5.2-High experiments cost approximately \$640. 


\section{Conclusion and Future Work}

We developed a novel approach for improving the performance of an LLM agent, based on automatically synthesizing a code harness.
Currently we generate a separate harness for each environment (game). In the future, we would like to distill the resulting domain specific experts (agents) back into the base LLM, so that the whole system becomes recursively self-improving.
We also hope to  explore building up a library of reusable harnesses, and to
apply our method to more challenging multimodal games,
such as Craftax \footnote{https://github.com/MichaelTMatthews/Craftax} 
and Terra Nova\footnote{https://github.com/trevormcinroe/terra\_nova/}.

\eat{
for learning code-as-harness for LLM, which empowers a much smaller Gemini-2.5-Flash model to rival much larger ones such as Gemini-2.5-Pro and GPT-5.2-High on large collection of TextArena games. Next, we are extending this work into a life-long learning paradigm and hope to solve very long-horizon games 
such as Craftax \footnote{https://github.com/MichaelTMatthews/Craftax} and Terra Nova\footnote{https://github.com/trevormcinroe/terra\_nova/}.
}

\bibliography{autoharness}

@string{NIPS = {{NeurIPS}}}

@article{tang2024code,
  title={Code repair with llms gives an exploration-exploitation tradeoff},
  author={Tang, Hao and Hu, Keya and Zhou, Jin and Zhong, Si Cheng and Zheng, Wei-Long and Si, Xujie and Ellis, Kevin},
  journal=NIPS,
  volume={37},
  pages={117954--117996},
  year={2024}
}

@article{wang2023voyager,
  title={Voyager: An open-ended embodied agent with large language models},
  author={Wang, Guanzhi and Xie, Yuqi and Jiang, Yunfan and Mandlekar, Ajay and Xiao, Chaowei and Zhu, Yuke and Fan, Linxi and Anandkumar, Anima},
  journal={arXiv preprint arXiv:2305.16291},
  year={2023}
}

@inproceedings{liang2023code,
  title={Code as Policies: Language Model Programs for Embodied Control},
  author={Liang, Jacky and Huang, Wenlong and Xia, Fei and Xu, Peng and Hausman, Karol and Ichter, Brian and Florence, Pete and Zeng, Andy},
  booktitle={ICRA},
  pages={9493--9500},
  year={2023},
  organization={IEEE},
  doi={10.1109/ICRA48891.2023.10160591}
}

@inproceedings{ma2024eureka,
  title={Eureka: Human-Level Reward Design via Coding Large Language Models},
  author={Yecheng Jason Ma and William Liang and Guanzhi Wang and De-An Huang and Osbert Bastani and Dinesh Jayaraman and Yuke Zhu and Linxi Fan and Anima Anandkumar},
  booktitle={ICLR},
  year={2024}
}

@article{li2022competition,
  title={Competition-level code generation with alphacode},
  author={Li, Yujia and Choi, David and Chung, Junyoung and Kushman, Nate and Schrittwieser, Julian and Leblond, R{\'e}mi and Eccles, Tom and Keeling, James and Gimeno, Felix and Dal Lago, Agustin and others},
  journal={Science},
  volume={378},
  number={6624},
  pages={1092--1097},
  year={2022}
}

@article{shinn2023reflexion,
  title={Reflexion: Language agents with verbal reinforcement learning},
  author={Shinn, Noah and Cassano, Federico and Gopinath, Ashwin and Narasimhan, Karthik and Yao, Shunyu},
  journal=NIPS,
  volume={36},
  pages={8634--8652},
  year={2023}
}

@article{guertler2025textarena,
  title={TextArena},
  author={Guertler, Leon and Cheng, Bobby and Yu, Simon and Liu, Bo and Choshen, Leshem and Tan, Cheston},
  journal={arXiv:2504.11442},
  year={2025}
}

@article{huang2025winning,
  title={Winning gold at IMO 2025 with a model-agnostic verification-and-refinement pipeline},
  author={Huang, Yichen and Yang, Lin F},
  journal={arXiv:2507.15855},
  year={2025}
}

@article{chervonyi2025gold,
  title={Gold-medalist performance in solving olympiad geometry with Alphageometry2},
  author={Chervonyi, Yuri and Trinh, Trieu H and Ol{\v{s}}{\'a}k, Miroslav and Yang, Xiaomeng and Nguyen, Hoang H and Menegali, Marcelo and Jung, Junehyuk and Kim, Junsu and Verma, Vikas and Le, Quoc V and others},
  journal={JMLR},
  volume={26},
  number={241},
  pages={1--39},
  year={2025}
}

@article{petrov2025proof,
  title={Proof or bluff? evaluating LLMs on 2025 USA math olympiad},
  author={Petrov, Ivo and Dekoninck, Jasper and Baltadzhiev, Lyuben and Drencheva, Maria and Minchev, Kristian and Balunovi{\'c}, Mislav and Jovanovi{\'c}, Nikola and Vechev, Martin},
  journal={arXiv:2503.21934},
  year={2025}
}

@article{novikov2025alphaevolve,
  title={AlphaEvolve: A coding agent for scientific and algorithmic discovery},
  author={Novikov, Alexander and V{\~u}, Ng{\^a}n and Eisenberger, Marvin and Dupont, Emilien and Huang, Po-Sen and Wagner, Adam Zsolt and Shirobokov, Sergey and Kozlovskii, Borislav and Ruiz, Francisco JR and Mehrabian, Abbas and others},
  journal={arXiv:2506.13131},
  year={2025}
}

@article{yao2023tree,
  title={Tree of thoughts: Deliberate problem solving with large language models},
  author={Yao, Shunyu and Yu, Dian and Zhao, Jeffrey and Shafran, Izhak and Griffiths, Tom and Cao, Yuan and Narasimhan, Karthik},
  journal=NIPS,
  volume={36},
  pages={11809--11822},
  year={2023}
}

@article{wei2022chain,
  title={Chain-of-thought prompting elicits reasoning in large language models},
  author={Wei, Jason and Wang, Xuezhi and Schuurmans, Dale and Bosma, Maarten and Xia, Fei and Chi, Ed and Le, Quoc V and Zhou, Denny and others},
  journal=NIPS,
  volume={35},
  pages={24824--24837},
  year={2022}
}

@article{valmeekam2023planning,
  title={On the planning abilities of large language models-a critical investigation},
  author={Valmeekam, Karthik and Marquez, Matthew and Sreedharan, Sarath and Kambhampati, Subbarao},
  journal=NIPS,
  volume={36},
  pages={75993--76005},
  year={2023}
}

@article{lehrach2025code,
  title={Code World Models for General Game Playing},
  author={Lehrach, Wolfgang and Hennes, Daniel and Lazaro-Gredilla, Miguel and Lou, Xinghua and Wendelken, Carter and Li, Zun and Dedieu, Antoine and Grau-Moya, Jordi and Lanctot, Marc and Iscen, Atil and others},
  journal={arXiv:2510.04542},
  year={2025}
}

@misc{kaggle_game_arena_2025,
  author = {Kaggle},
  title = {Kaggle Game Arena: A Benchmarking Platform for AI Models},
  year = {2025},
  howpublished = {\url{https://www.kaggle.com/game-arena}}
}

@INPROCEEDINGS{Kokel2025,
  title     = "{ACPBench} Hard: Unrestrained Reasoning about Action, Change, and
               Planning",
  author    = "Kokel, Harsha and Katz, Michael and Srinivas, Kavitha and
               Sohrabi, Shirin",
  booktitle = "AAAI 2025 Workshop LM4Plan",
  month     =  feb,
  year      =  2025,
}

@INPROCEEDINGS{Valmeekam2023,
  title     = "On the Planning Abilities of Large Language Models - A Critical
               Investigation",
  author    = "Valmeekam, Karthik and Marquez, Matthew and Sreedharan, Sarath
               and Kambhampati, Subbarao",
  booktitle = nips,
  month     =  nov,
  year      =  2023,
}

@ARTICLE{Ruoss2024,
  title  = "{LMAct}: A Benchmark for In-Context Imitation Learning with Long
            Multimodal Demonstrations",
  author = "Ruoss, Anian and Pardo, Fabio and Chan, Harris and Li, Bonnie and
            Mnih, Volodymyr and Genewein, Tim",
  month  =  dec,
  year   =  2024,
  url    = "http://arxiv.org/abs/2412.01441"
}

@ARTICLE{Duan2024,
  title         = "{GTBench}: Uncovering the strategic reasoning limitations of
                   {LLMs} via game-theoretic evaluations",
  author        = "Duan, Jinhao and Zhang, Renming and Diffenderfer, James and
                   Kailkhura, Bhavya and Sun, Lichao and Stengel-Eskin, Elias
                   and Bansal, Mohit and Chen, Tianlong and Xu, Kaidi",
  journal       = "arXiv [cs.CL]",
  month         =  feb,
  year          =  2024,
  archivePrefix = "arXiv",
  primaryClass  = "cs.CL"
}
\bibliographystyle{autoharness}

\newpage
\appendix

\section{TextArena games}
\label{sec:games}

\eat{
\subsection{Full List of 1-Player Games}
\begin{lstlisting}
    '2048-v0'
    'Bandit-v0'
    'Blackjack-v0'
    'Cryptarithm-v0'
    'FifteenPuzzle-v0'
    'FrozenLake-v0'
    'GuessTheNumber-v0'
    'LightsOut-v0'
    'Mastermind-v0'
    'Minesweeper-v0'
    'PegJump-v0'
    'RushHour-v0'
    'Secretary-v0'
    'Sokoban-v0'
    'Sudoku-v0'
    'TowerOfHanoi-v0'
\end{lstlisting}

\subsection{Full List of 2-Player Games}
\begin{lstlisting}
    'LinesOfAction-v0'
    'Stratego-v0'
    'NewRecruit-v0'
    'Chess-v0'
    'Alquerque-v0'
    'Tak-v0'
    'LiarsDice-v0'
    'Golf-v0'
    'Chopsticks-v0'
    'Checkers-v0'
    'Breakthrough-v0'
    'Crusade-v0'
    'SpiteAndMalice-v0'
    'UltimateTicTacToe-v0'
    'GermanWhist-v0'
    'Othello-v0'
\end{lstlisting}

\subsection{FULL LIST OF GAMES WITH “AVAILABLE MOVES” REMOVED}

\begin{lstlisting}
    'Chess-v0'
    'Crusade-v0'
    'Golf-v0'
    'Othello-v0'
    'SpiteAndMalice-v0'
    'FifteenPuzzle-v0'
    'FrozenLake-v0'
    'Sokoban-v0'
\end{lstlisting}

}

\subsection{List of all 145 games}

\begin{longtable}{|l|l|r|r|r|}
\caption{List of all 145 TextArena games, with accuracy of learned harness, and number of LLM calls needed to achieve this. The 32 games used for end-to-end agent eval are marked with *.} \label{tab:long} \\

\hline \multicolumn{1}{|c|}{\textbf{Index}} & \multicolumn{1}{c|}{\textbf{Game}} & \multicolumn{1}{c|}{\textbf{\# Players}} & \multicolumn{1}{c|}{\textbf{\# Learning Steps}} & \multicolumn{1}{c|}{\textbf{Legal Action Rate}} \\ \hline 

\endfirsthead

\multicolumn{5}{c}%
{{\bfseries \tablename\ \thetable{} -- continued from previous page}} \\
\hline \multicolumn{1}{|c|}{\textbf{Index}} & \multicolumn{1}{c|}{\textbf{Game}} & \multicolumn{1}{c|}{\textbf{\# Players}} & \multicolumn{1}{c|}{\textbf{\# Learning Steps}} & \multicolumn{1}{c|}{\textbf{Legal Action Rate}} \\ \hline 
\endhead

\hline \multicolumn{5}{|r|}{{Continued on next page}} \\ \hline
\endfoot

\hline \hline
\endlastfoot

0 & 2048-v0* & 1 & 27 & 1.0 \\
1 & 2048-v0-easy & 1 & 4 & 1.0 \\
2 & 2048-v0-extreme & 1 & 44 & 1.0 \\
3 & 2048-v0-hard & 1 & 47 & 1.0 \\
4 & 2048-v0-mega-easy & 1 & 31 & 1.0 \\
5 & 2048-v0-super-easy & 1 & 6 & 1.0 \\
6 & 2048-v0-ultra-easy & 1 & 2 & 1.0 \\
7 & 2048-v0-very-easy & 1 & 57 & 1.0 \\
8 & 2048-v0-very-hard & 1 & 7 & 1.0 \\
9 & Alquerque-v0* & 2 & 4 & 1.0 \\
10 & Bandit-v0* & 1 & 2 & 1.0 \\
11 & Bandit-v0-hard & 1 & 1 & 1.0 \\
12 & Battleship-v0 & 2 & 4 & 1.0 \\
13 & Battleship-v0-extreme & 2 & 32 & 1.0 \\
14 & Battleship-v0-large & 2 & 9 & 1.0 \\
15 & Battleship-v0-standard & 2 & 6 & 1.0 \\
16 & Blackjack-v0* & 1 & 2 & 1.0 \\
17 & Blackjack-v0-long & 1 & 1 & 1.0 \\
18 & Breakthrough-v0* & 2 & 2 & 1.0 \\
19 & Breakthrough-v0-blind & 2 & 20 & 1.0 \\
20 & Breakthrough-v0-large & 2 & 9 & 1.0 \\
21 & Breakthrough-v0-long & 2 & 7 & 1.0 \\
22 & Breakthrough-v0-small & 2 & 136 & 1.0 \\
23 & Breakthrough-v0-tiny & 2 & 5 & 1.0 \\
24 & Briscola-v0 & 2 & 2 & 1.0 \\
25 & Checkers-v0* & 2 & 7 & 1.0 \\
26 & Checkers-v0-long & 2 & 3 & 1.0 \\
27 & Chess-v0* & 2 & 64 & 1.0 \\
28 & Chess-v0-blind & 2 & 19 & 1.0 \\
29 & Chess-v0-long & 2 & 16 & 1.0 \\
30 & Chopsticks-v0* & 2 & 15 & 1.0 \\
31 & Chopsticks-v0-long & 2 & 7 & 1.0 \\
32 & Chopsticks-v0-medium & 2 & 15 & 1.0 \\
33 & ColonelBlotto-v0 & 2 & 1 & 1.0 \\
34 & ColonelBlotto-v0-extreme & 2 & 1 & 1.0 \\
35 & ColonelBlotto-v0-large & 2 & 1 & 1.0 \\
36 & ColonelBlotto-v0-small & 2 & 1 & 1.0 \\
37 & ConnectFour-v0 & 2 & 10 & 1.0 \\
38 & ConnectFour-v0-blind & 2 & 2 & 1.0 \\
39 & ConnectFour-v0-large & 2 & 1 & 1.0 \\
40 & Crusade-v0* & 2 & 4 & 1.0 \\
41 & Cryptarithm-v0* & 1 & 45 & 1.0 \\
42 & FifteenPuzzle-v0* & 1 & 3 & 1.0 \\
43 & FrozenLake-v0* & 1 & 19 & 1.0 \\
44 & FrozenLake-v0-hardcore & 1 & 4 & 1.0 \\
45 & FrozenLake-v0-random & 1 & 22 & 1.0 \\
46 & GameOfPureStrategy-v0 & 2 & 3 & 1.0 \\
47 & GermanWhist-v0* & 2 & 43 & 1.0 \\
48 & Golf-v0* & 2 & 8 & 1.0 \\
49 & Golf-v0-medium & 2 & 9 & 1.0 \\
50 & GuessTheNumber-v0* & 1 & 2 & 1.0 \\
51 & GuessTheNumber-v0-hardcore & 1 & 2 & 1.0 \\
52 & HighSociety-v0 & 2 & 3 & 1.0 \\
53 & IndianPoker-v0 & 2 & 11 & 1.0 \\
54 & IndianPoker-v0-extreme & 2 & 2 & 1.0 \\
55 & IndianPoker-v0-long & 2 & 26 & 1.0 \\
56 & IndianPoker-v0-medium & 2 & 7 & 1.0 \\
57 & IndianPoker-v0-short & 2 & 2 & 1.0 \\
58 & IteratedMatchingPennies-v0 & 2 & 1 & 1.0 \\
59 & IteratedRockPaperScissors-v0 & 2 & 1 & 1.0 \\
60 & IteratedTwoThirdsAverage-v0 & 2 & 1 & 1.0 \\
61 & KuhnPoker-v0 & 2 & 5 & 1.0 \\
62 & KuhnPoker-v0-extreme & 2 & 3 & 1.0 \\
63 & KuhnPoker-v0-long & 2 & 2 & 1.0 \\
64 & KuhnPoker-v0-medium & 2 & 2 & 1.0 \\
65 & KuhnPoker-v0-short & 2 & 3 & 1.0 \\
66 & LiarsDice-v0* & 2 & 4 & 1.0 \\
67 & LiarsDice-v0-large & 2 & 6 & 1.0 \\
68 & LiarsDice-v0-small & 2 & 5 & 1.0 \\
69 & LightsOut-v0* & 1 & 1 & 1.0 \\
70 & LinesOfAction-v0* & 2 & 23 & 1.0 \\
71 & Mastermind-v0* & 1 & 2 & 1.0 \\
72 & Mastermind-v0-extreme & 1 & 1 & 1.0 \\
73 & Mastermind-v0-hard & 1 & 2 & 1.0 \\
74 & MemoryGame-v0 & 2 & 3 & 1.0 \\
75 & MemoryGame-v0-hard & 2 & 2 & 1.0 \\
76 & MemoryGame-v0-medium & 2 & 2 & 1.0 \\
77 & Minesweeper-v0* & 1 & 11 & 1.0 \\
78 & Minesweeper-v0-hard & 1 & 6 & 1.0 \\
79 & Minesweeper-v0-medium & 1 & 10 & 1.0 \\
80 & Minesweeper-v0-small & 1 & 2 & 1.0 \\
81 & NewRecruit-v0* & 2 & 2 & 1.0 \\
82 & Nim-v0 & 2 & 1 & 1.0 \\
83 & Nim-v0-large & 2 & 2 & 1.0 \\
84 & Nim-v0-medium & 2 & 2 & 1.0 \\
85 & Othello-v0* & 2 & 62 & 1.0 \\
86 & Othello-v0-big & 2 & 2 & 1.0 \\
87 & Othello-v0-hard & 2 & 30 & 1.0 \\
88 & Othello-v0-huge & 2 & 12 & 1.0 \\
89 & Othello-v0-small & 2 & 5 & 1.0 \\
90 & Othello-v0-tiny & 2 & 13 & 1.0 \\
91 & PegJump-v0* & 1 & 1 & 1.0 \\
92 & PigDice-v0 & 2 & 1 & 1.0 \\
93 & PigDice-v0-100 & 2 & 1 & 1.0 \\
94 & PigDice-v0-150 & 2 & 1 & 1.0 \\
95 & PigDice-v0-200 & 2 & 1 & 1.0 \\
96 & PigDice-v0-250 & 2 & 1 & 1.0 \\
97 & PigDice-v0-300 & 2 & 1 & 1.0 \\
98 & PigDice-v0-350 & 2 & 1 & 1.0 \\
99 & PigDice-v0-400 & 2 & 1 & 1.0 \\
100 & PigDice-v0-450 & 2 & 1 & 1.0 \\
101 & PigDice-v0-50 & 2 & 1 & 1.0 \\
102 & PigDice-v0-500 & 2 & 1 & 1.0 \\
103 & PigDice-v0-long & 2 & 1 & 1.0 \\
104 & PigDice-v0-short & 2 & 1 & 1.0 \\
105 & Poker-v0 & 2 & 17 & 1.0 \\
106 & Poker-v0-extreme & 2 & 7 & 1.0 \\
107 & Poker-v0-long & 2 & 5 & 1.0 \\
108 & Poker-v0-small & 2 & 29 & 1.0 \\
109 & QuantumTicTacToe-v0 & 2 & 12 & 1.0 \\
110 & ReverseTicTacToe-v0 & 2 & 3 & 1.0 \\
111 & RushHour-v0* & 1 & 3 & 1.0 \\
112 & SantoriniBaseFixed-v0 & 2 & 30 & 1.0 \\
113 & Secretary-v0* & 1 & 1 & 1.0 \\
114 & Secretary-v0-long & 1 & 1 & 1.0 \\
115 & SimpleTak-v0 & 2 & 4 & 1.0 \\
116 & SimpleTak-v0-extreme & 2 & 8 & 1.0 \\
117 & SimpleTak-v0-large & 2 & 12 & 1.0 \\
118 & SimpleTak-v0-medium & 2 & 5 & 1.0 \\
119 & Snake-v0 & 2 & 1 & 1.0 \\
120 & Snake-v0-large & 2 & 1 & 1.0 \\
121 & Snake-v0-standard & 2 & 1 & 1.0 \\
122 & Sokoban-v0* & 1 & 5 & 1.0 \\
123 & Sokoban-v0-medium & 1 & 1 & 1.0 \\
124 & SpiteAndMalice-v0* & 2 & 33 & 1.0 \\
125 & Stratego-v0* & 2 & 23 & 1.0 \\
126 & Sudoku-v0* & 1 & 5 & 1.0 \\
127 & Sudoku-v0-easy & 1 & 5 & 1.0 \\
128 & Sudoku-v0-hard & 1 & 9 & 1.0 \\
129 & Sudoku-v0-medium & 1 & 4 & 1.0 \\
130 & Sudoku-v0-very-easy & 1 & 4 & 1.0 \\
131 & Surround-v0 & 2 & 1 & 1.0 \\
132 & Surround-v0-large & 2 & 1 & 1.0 \\
133 & Surround-v0-standard & 2 & 1 & 1.0 \\
134 & Tak-v0* & 2 & 21 & 1.0 \\
135 & Tak-v0-hard & 2 & 53 & 1.0 \\
136 & Tak-v0-medium & 2 & 6 & 1.0 \\
137 & TicTacToe-v0 & 2 & 4 & 1.0 \\
138 & TowerOfHanoi-v0* & 1 & 7 & 1.0 \\
139 & TowerOfHanoi-v0-extreme & 1 & 44 & 1.0 \\
140 & TowerOfHanoi-v0-hard & 1 & 7 & 1.0 \\
141 & TowerOfHanoi-v0-hardcore & 1 & 2 & 1.0 \\
142 & TowerOfHanoi-v0-medium & 1 & 7 & 1.0 \\
143 & UltimateTicTacToe-v0* & 2 & 13 & 1.0 \\
144 & WildTicTacToe-v0 & 2 & 10 & 1.0 \\

\end{longtable}

\subsection{Per-game reward}

\begin{figure}[H]
    \centering
    \includegraphics[width=0.99\textwidth]{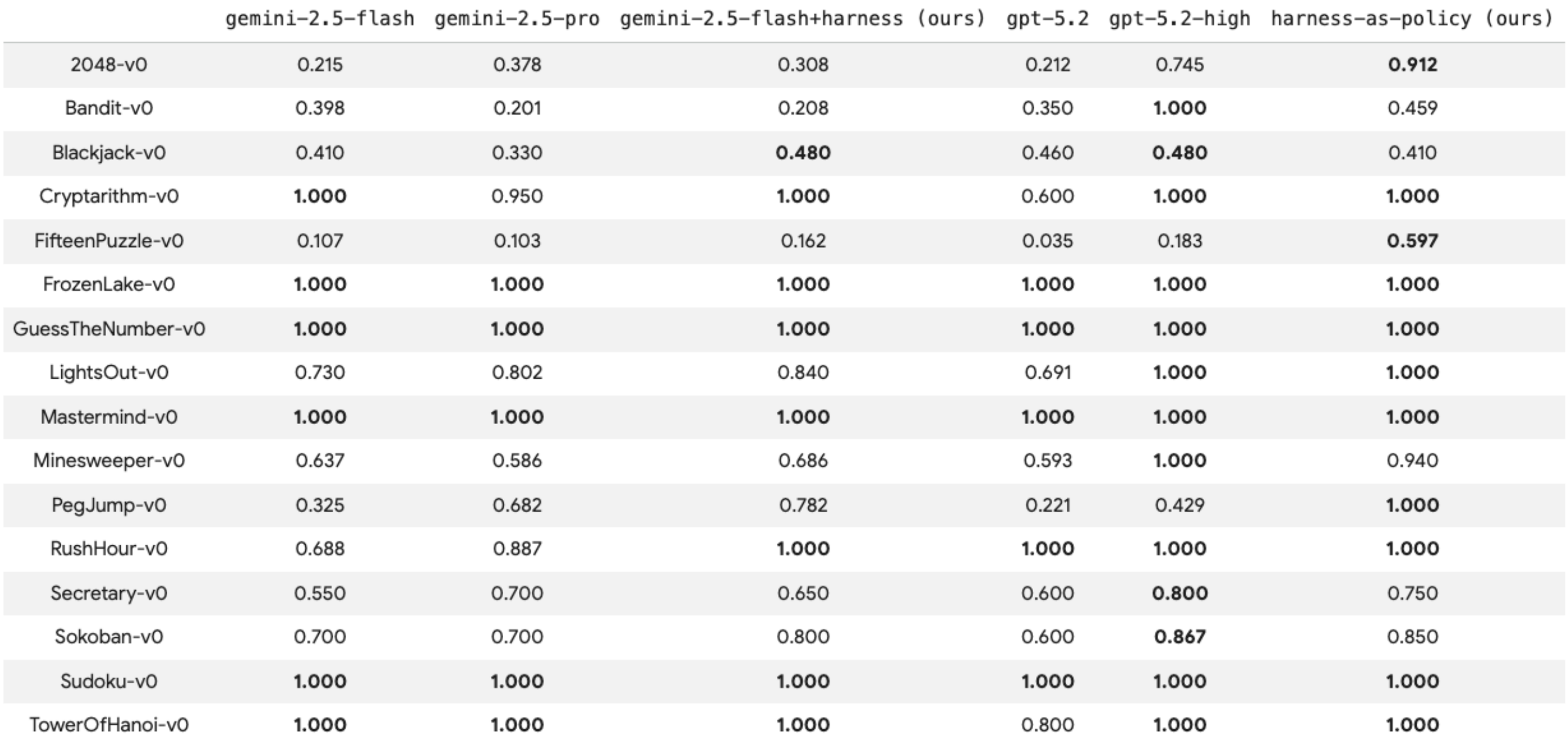}
    \caption{TextArena 1P per-game reward.}
    \label{fig:1p-breakdown-reward} 
\end{figure}

\subsection{Per-game Legal Action Rate}
\begin{figure}[H]
    \centering
    \includegraphics[width=0.99\textwidth]{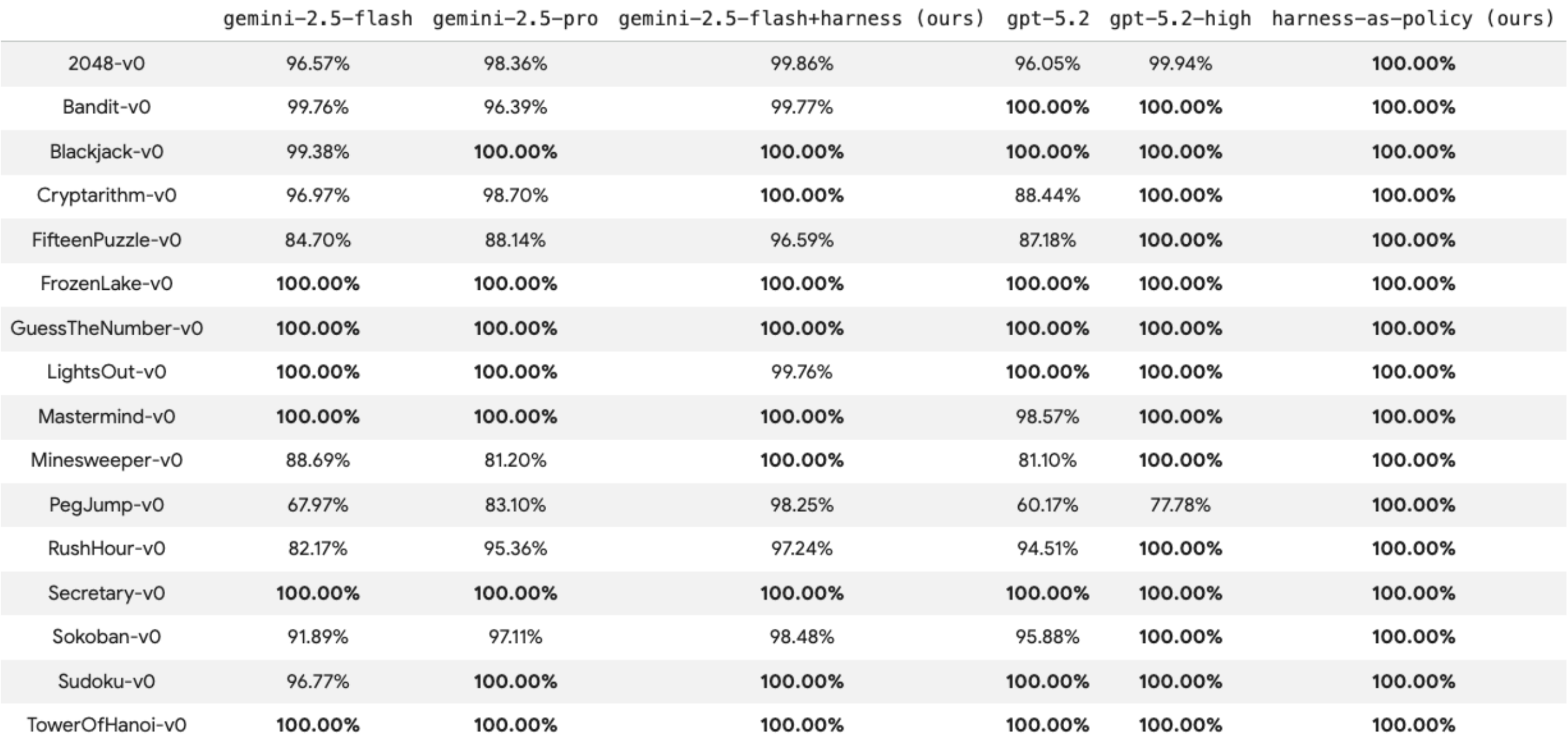}
    \caption{TextArena 1P per-game legal action success rate.}
    \label{fig:1p-breakdown-legal-action-rate} 
\end{figure}

\subsection{Example game: Chess-v0}
\label{sec:remove-legal-chess}

In this section, we illustrate how we remove the list of legal actions from the observation.

\subsubsection{Original Chess-v0 observation}
\begin{lstlisting}
[GAME] You are playing White in a game of Chess.
 Make your moves in UCI format enclosed in square brackets (e.g., [e2e4]).
[GAME] Current board:
   +-----------------+
 8 | r n b q k b n r |
 7 | p p p p p p p p |
 6 | . . . . . . . . |
 5 | . . . . . . . . |
 4 | . . . . . . . . |
 3 | . . . . . . . . |
 2 | P P P P P P P P |
 1 | R N B Q K B N R |
   +-----------------+
    a b c d e f g h 
Valid moves: [g1h3], [g1f3], [b1c3], [b1a3], [h2h3], [g2g3], [f2f3], [e2e3], [d2d3], [c2c3], [b2b3], [a2a3], [h2h4], [g2g4], [f2f4], [e2e4], [d2d4], [c2c4], [b2b4], [a2a4]
\end{lstlisting}

\subsection{Modified Chess-v0 observation with ``valid moves'' removed}
\begin{lstlisting}
[GAME] You are playing White in a game of Chess.
 Make your moves in UCI format enclosed in square brackets (e.g., [e2e4]).
[GAME] Current board:
   +-----------------+
 8 | r n b q k b n r |
 7 | p p p p p p p p |
 6 | . . . . . . . . |
 5 | . . . . . . . . |
 4 | . . . . . . . . |
 3 | . . . . . . . . |
 2 | P P P P P P P P |
 1 | R N B Q K B N R |
   +-----------------+
    a b c d e f g h 
\end{lstlisting}

\section{Prompts}

\subsection{LLM-as-policy prompt}

\begin{lstlisting}
You are an expert, logical, and strategic AI game player. Your task is to analyze the following game information and determine the single best move to make.

Read the game rules, your player role, the current game state, and all available moves carefully. Your objective is to play optimally to maximize your chances of winning the game.

You are now player {player_id}.

The game information is as follows:
{observation}

**YOUR TASK:**

You must now analyze the situation and provide your move. Follow these two steps precisely.

**Step 1: Think**
First, provide your step-by-step reasoning. Analyze the current game state, your goal, and the available moves. Evaluate the pros and cons of the most promising options and explain why you are selecting your final move.

**Step 2: Move**
After your thinking block, provide *only* the single best move you have chosen. The move must be one of the valid moves listed in the game information.

Enclose your final move in `<move></move>` tags. Do not add any other text, explanation, or punctuation after the closing `</move>` tag.

Example of a correct response format:
<move>
[Your chosen move]
</move>
\end{lstlisting}

\subsection{Code Refinement Prompt}

\begin{lstlisting}
You are a python programmer with expertise in text games.

You are given a text game with the following name: {name}

Here is a description of the game.
{description}

Here is a description of the action space of the game.
{action_space}

You are observing the following game boards as text with error feedback.
{tasks_with_feedback}

Your task is to write or refine the following python functions.
```python
{code}
```

Make sure to follow these function signatures.
```python
{code_signatures}
```

Make sure to follow these instructions.

* Think step by step about the code, the game boards and the error feedback.

* Reason about each action through the game board and write down critical failure steps.

* Reason about code refinements that can help fix the failure steps.

* Reason about the entire sequence of actions and write down the progress of the game as a value between 0 and 1.

* Reason about code refinements that can help improve the game progress.

* Reason about code refinements that can avoid running in loops.

* Write down your thoughts before writing the code.

* Make sure to follow the given function signatures.

* Make sure the new code can satisfy all the observed game boards.

* Make sure the new code can fix all the current errors.

* Make sure to only produce code that is safe to execute.

* Make sure the code is concise and precise.

* If necessary, randomply sample one of the best legal actions and return it as the proposed action.

* Do not use any the try-except blocks.

* Write your functions in a python code block enclosed in ```python
\end{lstlisting}

\section{Harness Function Signatures}

\subsection{Code-as-action-verifier}

\begin{lstlisting}[language=Python,showstringspaces=false]
def propose_action(board: str) -> str:
  """Propose a valid random action given the game board as text

  Args:
      board (str): Game board as text.

  Returns:
      str: A valid random action as string.

  Raises:
      Exception: If fail to propose a valid random action.
  """
  raise NotImplementedError()

def is_legal_action(board: str, action: str) -> bool:
  """Check if an action string is valid given the game board as text

  Args:
      board (str): Game board as text.
      action (str): Input action as string.

  Returns:
      bool: If the input action string is valid.

  Raises:
      Exception: If fail to check if the action string is valid.
  """
  raise NotImplementedError()
\end{lstlisting}

\subsection{Harness-as-policy}

We use the same function signatures as above, except the docstring of \texttt{propose\_action()}:
\begin{lstlisting}
Propose one of the best legal actions given the game board as text such that the final reward is maximized.
\end{lstlisting}

\section{Sample Harness Code Snippets}
\label{sec:snippets}

\subsection{Minesweeper-v0}

The \texttt{propose\_action()} code snippet for Minesweeper-v0 breaks down the strategy by checking the first move, finding guaranteed safe cells by logic deduction, and applying probabilistic heuristic for best guesses. Not that not the whole code harness is shown here.

\begin{lstlisting}[language=Python,showstringspaces=false]
def propose_action(board: str) -> str:
  """Propose one of the best legal actions given the game board as text such that the final reward is maximized.

  Args:
      board (str): Game board as text.

  Returns:
      str: A string representing one of the best legal actions.

  Raises:
      Exception: If fail to propose any legal action.
  """
  grid = parse_board_to_grid(board)
  if not grid:
    raise Exception("Failed to parse the board or board is empty, cannot propose an action.")

  num_rows, num_cols = get_board_dimensions(grid)
  if num_rows == 0 or num_cols == 0:
    raise Exception("Board dimensions are zero, cannot propose an action.")

  # Check if this is the very first move (all cells are unrevealed)
  all_cells_unrevealed = True
  for r_check in range(num_rows):
    for c_check in range(num_cols):
      if grid[r_check][c_check] != '.':
        all_cells_unrevealed = False
        break
    if not all_cells_unrevealed:
      break

  # Strategy 1: First move (if all cells are unrevealed)
  # Pick a central cell to maximize the initial revealed area for a safe start.
  if all_cells_unrevealed:
    # For an 8x8 board, (3,3) is a common safe starting point.
    first_move_row = num_rows // 2 - (1 if num_rows % 2 == 0 and num_rows // 2 > 0 else 0)
    first_move_col = num_cols // 2 - (1 if num_cols % 2 == 0 and num_cols // 2 > 0 else 0)
    return f"[{first_move_row} {first_move_col}]"

  # Strategy 2: Logic Deduction for Guaranteed Safe Cells and Mines
  # board_mines stores: True for known mine, False for known safe, None for unknown.
  board_mines: list[list[bool | None]] = [[None for _ in range(num_cols)] for _ in range(num_rows)]
  
  # Initialize board_mines based on currently revealed cells
  for r in range(num_rows):
      for c in range(num_cols):
          if grid[r][c].isdigit():
              # Revealed cells are definitely not mines.
              board_mines[r][c] = False 
          # Unrevealed cells '.' are initially None (unknown)

  safe_to_reveal: list[tuple[int, int]] = [] # Cells that are deduced to be safe to click
  safe_cells_set = set() # Use a set to track already added safe cells to avoid duplicates

  # Propagate deductions multiple times until no new deductions are made
  max_deduction_iterations = num_rows * num_cols # Safety limit to ensure termination
  for _ in range(max_deduction_iterations): 
      new_deductions_made_this_iteration = False # Track deductions per iteration

      # --- Simple Deduction Rules (Rule A & B) ---
      for r in range(num_rows):
          for c in range(num_cols):
              if grid[r][c].isdigit(): 
                  N = int(grid[r][c])
                  
                  unrevealed_unknown_neighbors = [] # Neighbors that are '.' and NOT yet marked as mine/safe (None in board_mines)
                  known_mine_neighbors_count = 0
                  
                  for dr in [-1, 0, 1]:
                      for dc in [-1, 0, 1]:
                          if dr == 0 and dc == 0:
                              continue
                          nr, nc = r + dr, c + dc
                          
                          if 0 <= nr < num_rows and 0 <= nc < num_cols:
                              if board_mines[nr][nc] is True:
                                  known_mine_neighbors_count += 1
                              elif grid[nr][nc] == '.' and board_mines[nr][nc] is None:
                                  unrevealed_unknown_neighbors.append((nr, nc))
                  
                  num_unrevealed_and_unknown = len(unrevealed_unknown_neighbors)
                  mines_to_deduce = N - known_mine_neighbors_count

                  # Rule A: Deduce Mines
                  # If (clue N - known_mine_neighbors_count) equals the number of unrevealed and unknown neighbors,
                  # then all these unrevealed and unknown neighbors must be mines.
                  if mines_to_deduce > 0 and mines_to_deduce == num_unrevealed_and_unknown:
                      for (ur, uc) in unrevealed_unknown_neighbors:
                          if board_mines[ur][uc] is None: # Only mark if not already marked
                              board_mines[ur][uc] = True
                              new_deductions_made_this_iteration = True
                  
                  # Rule B: Deduce Safe Cells
                  # If clue N equals the number of known mine neighbors,
                  # then all other unrevealed and unknown neighbors must be safe.
                  elif mines_to_deduce == 0 and num_unrevealed_and_unknown > 0:
                      for (ur, uc) in unrevealed_unknown_neighbors:
                          if board_mines[ur][uc] is None: # Only mark if not already marked
                              board_mines[ur][uc] = False # Mark as safe
                              if (ur, uc) not in safe_cells_set: # Add to safe list only if new
                                  safe_to_reveal.append((ur, uc))
                                  safe_cells_set.add((ur, uc))
                              new_deductions_made_this_iteration = True

      # --- Advanced Deduction Rules (Subset Rule) ---
      # Collect all clue constraints that have unknown neighbors
      clue_constraints = [] # List of (mines_needed, set_of_unknown_neighbors)
      for r_clue in range(num_rows):
          for c_clue in range(num_cols):
              if grid[r_clue][c_clue].isdigit():
                  N_clue = int(grid[r_clue][c_clue])
                  unknown_neighbors_set = set()
                  known_mines_around_clue = 0

                  for dr_clue in [-1, 0, 1]:
                      for dc_clue in [-1, 0, 1]:
                          if dr_clue == 0 and dc_clue == 0:
                              continue
                          nr_clue, nc_clue = r_clue + dr_clue, c_clue + dc_clue
                          if 0 <= nr_clue < num_rows and 0 <= nc_clue < num_cols:
                              if board_mines[nr_clue][nc_clue] is True:
                                  known_mines_around_clue += 1
                              elif grid[nr_clue][nc_clue] == '.' and board_mines[nr_clue][nc_clue] is None:
                                  unknown_neighbors_set.add((nr_clue, nc_clue))
                  
                  mines_needed = N_clue - known_mines_around_clue
                  # Only add valid constraints where mines_needed is positive and there are unknown cells to consider.
                  # If mines_needed <= 0, Rule B (safe cells) would have already covered it or it means too many mines are already identified.
                  if mines_needed > 0 and unknown_neighbors_set: 
                      clue_constraints.append((mines_needed, unknown_neighbors_set))

      # Apply subset rule to all pairs of clue constraints
      for i in range(len(clue_constraints)):
          for j in range(len(clue_constraints)):
              if i == j: # Don't compare a constraint with itself
                  continue
              
              nm1, s1 = clue_constraints[i] # mines_needed_1, set_of_unknown_neighbors_1
              nm2, s2 = clue_constraints[j] # mines_needed_2, set_of_unknown_neighbors_2

              if s1.issubset(s2) and s1 != s2: # If s1 is a proper subset of s2
                  s_diff = s2 - s1 # Cells in s2 but not in s1
                  nm_diff = nm2 - nm1 # Difference in needed mines

                  if nm_diff == 0 and s_diff: # If difference in mines is 0, then s_diff cells must be safe
                      for (sr, sc) in s_diff:
                          if board_mines[sr][sc] is None:
                              board_mines[sr][sc] = False
                              if (sr, sc) not in safe_cells_set:
                                  safe_to_reveal.append((sr, sc))
                                  safe_cells_set.add((sr, sc))
                              new_deductions_made_this_iteration = True
                  elif nm_diff == len(s_diff) and s_diff: # If difference in mines equals number of cells in s_diff, they are all mines
                      for (sr, sc) in s_diff:
                          if board_mines[sr][sc] is None:
                              board_mines[sr][sc] = True
                              new_deductions_made_this_iteration = True
      
      if not new_deductions_made_this_iteration:
          break # No new deductions this iteration (neither simple nor advanced), stop propagating

  # If guaranteed safe cells are found, choose one randomly.
  if safe_to_reveal:
    chosen_move = random.choice(safe_to_reveal)
    return f"[{chosen_move[0]} {chosen_move[1]}]"

  # Strategy 3: Probabilistic Heuristic for Best Guess (when no guaranteed safe cells are found)
  potential_moves_with_risks = [] # Stores (risk_score, row, col)

  total_unrevealed_unknown_dots = 0
  identified_mines_count = 0
  for r in range(num_rows):
      for c in range(num_cols):
          if board_mines[r][c] is True:
              identified_mines_count += 1
          elif grid[r][c] == '.' and board_mines[r][c] is None:
              total_unrevealed_unknown_dots += 1

  # Assuming a standard 8x8 Minesweeper board has 10 mines. This is a common setup.
  total_mines_on_board = 10 
  
  global_mine_prob = 0.0
  if total_unrevealed_unknown_dots > 0:
      remaining_mines_to_place = max(0, total_mines_on_board - identified_mines_count)
      global_mine_prob = remaining_mines_to_place / total_unrevealed_unknown_dots
      # Clamp global probability between 0 and 1
      global_mine_prob = max(0.0, min(1.0, global_mine_prob))

  # Iterate over all unrevealed cells that are not known mines to calculate risk scores
  for r in range(num_rows):
      for c in range(num_cols):
          # Only consider truly unknown cells ('.') for guessing that are not marked as mines
          if grid[r][c] == '.' and board_mines[r][c] is None:
              current_cell_risk_sum = 0.0
              num_adjacent_clues_influencing = 0 

              # Analyze neighbors of the candidate cell (r, c)
              for dr_adj in [-1, 0, 1]:
                  for dc_adj in [-1, 0, 1]:
                      if dr_adj == 0 and dc_adj == 0:
                          continue
                      nr_adj, nc_adj = r + dr_adj, c + dc_adj
                      
                      if 0 <= nr_adj < num_rows and 0 <= nc_adj < num_cols:
                          if grid[nr_adj][nc_adj].isdigit(): # If adjacent to a revealed clue
                              num_adjacent_clues_influencing += 1
                              N_clue = int(grid[nr_adj][nc_adj])
                              
                              mines_around_clue = 0
                              unknown_around_clue_for_clue = 0 # Unknown neighbors specifically for THIS clue's context
                              
                              for dr_sub in [-1, 0, 1]:
                                  for dc_sub in [-1, 0, 1]:
                                      if dr_sub == 0 and dc_sub == 0:
                                          continue
                                      snr, snc = nr_adj + dr_sub, nc_adj + dc_sub

                                      if 0 <= snr < num_rows and 0 <= snc < num_cols:
                                          if board_mines[snr][snc] is True:
                                              mines_around_clue += 1
                                          elif grid[snr][snc] == '.' and board_mines[snr][snc] is None:
                                              unknown_around_clue_for_clue += 1
                              
                              remaining_mines_for_clue = N_clue - mines_around_clue
                              
                              if unknown_around_clue_for_clue > 0:
                                  # Probability of an unknown cell being a mine given this clue
                                  prob_from_clue = max(0.0, min(1.0, remaining_mines_for_clue / unknown_around_clue_for_clue))
                                  current_cell_risk_sum += prob_from_clue

              if num_adjacent_clues_influencing > 0: 
                  # Average the probabilities from all influencing clues
                  current_risk_score = current_cell_risk_sum / num_adjacent_clues_influencing
              else: # If the candidate cell is isolated from any revealed numbers, use global probability
                  current_risk_score = global_mine_prob 
              
              potential_moves_with_risks.append((current_risk_score, r, c))
              
  # Select the move(s) with the minimum risk score
  if potential_moves_with_risks:
      # Find the minimum risk score among all potential moves
      min_risk_score = float('inf')
      for risk, _, _ in potential_moves_with_risks:
          min_risk_score = min(min_risk_score, risk)
      
      # Collect all moves that have this minimum risk score
      best_moves = [(r, c) for risk, r, c in potential_moves_with_risks if risk == min_risk_score]

      if best_moves:
          # Randomly choose one of the best (lowest risk) moves
          chosen_move = random.choice(best_moves)
          return f"[{chosen_move[0]} {chosen_move[1]}]"

  # Final Fallback: If no moves passed the filters (e.g., all remaining cells are considered mines,
  # or no logically safe/guessable moves were found from the current state), 
  # but there are still unrevealed cells, pick one randomly from truly unknown cells.
  unrevealed_cells_remaining = []
  for r in range(num_rows):
      for c in range(num_cols):
          if grid[r][c] == '.':
              # Only pick truly unrevealed cells that are not yet marked as mines (is None)
              if board_mines[r][c] is None:
                  unrevealed_cells_remaining.append((r,c))
  
  if unrevealed_cells_remaining:
      chosen_move = random.choice(unrevealed_cells_remaining)
      return f"[{chosen_move[0]} {chosen_move[1]}]"

  raise Exception("No legal actions can be proposed. All non-mine cells might be revealed or no safe/guessable moves found.")
\end{lstlisting}

\subsection{Chess-v0}

Interesting code snippets for Chess-v0 including Universal Chess Interface (UCI) parsing and formatting, piece localizing and attack checking. Note that not the whole code harness is shown here.

\begin{lstlisting}
def _to_uci_coord(row: int, col: int) -> str:
    """Converts 0-indexed grid coordinates (row, col) to UCI string (e.g., 'e2')."""
    file_char = chr(ord('a') + col)
    rank_char = str(8 - row) # Grid row 0 is rank 8, grid row 7 is rank 1
    return file_char + rank_char

def _from_uci_coord(coord_str: str) -> tuple[int, int] | None:
    """Converts UCI string (e.g., 'e2') to 0-indexed grid coordinates (row, col). Returns None on invalid input."""
    if not (len(coord_str) == 2 and 'a' <= coord_str[0] <= 'h' and '1' <= coord_str[1] <= '8'):
        return None
    col = ord(coord_str[0]) - ord('a')
    row = 8 - int(coord_str[1]) # Rank 8 is grid row 0, rank 1 is grid row 7
    return row, col

def _find_king(grid: list[list[str]], king_color: str) -> tuple[int, int] | None:
    """Finds the coordinates of the king of the specified color."""
    for r in range(8):
        for c in range(8):
            piece = grid[r][c]
            if (king_color == 'w' and piece == 'K') or \
               (king_color == 'b' and piece == 'k'):
                return r, c
    return None # King not found (should ideally not happen in a valid game state)

def _is_square_attacked(grid: list[list[str]], r: int, c: int, by_white: bool) -> bool:
    """Checks if square (r, c) is attacked by any piece of the color 'by_white' (True for White, False for Black)."""
    
    # Helper to check if a piece at (pr, pc) is of the attacking_color
    def is_attacker(piece_sym: str, is_white_attacker: bool) -> bool:
        if piece_sym == '.': return False
        return (is_white_attacker and piece_sym.isupper()) or \
               (not is_white_attacker and piece_sym.islower())

    # 1. Pawn attacks (diagonal 1 step)
    # If checking attack *by* White, White pawns attack "up" (decreasing row index). So a white pawn attacking (r,c) would be at (r+1, c-1) or (r+1, c+1).
    # If checking attack *by* Black, Black pawns attack "down" (increasing row index). So a black pawn attacking (r,c) would be at (r-1, c-1) or (r-1, c+1).
    pawn_attacker_dr_from_target = 1 if by_white else -1 
    for dc_pawn in [-1, 1]:
        pr, pc = r + pawn_attacker_dr_from_target, c + dc_pawn
        if 0 <= pr < 8 and 0 <= pc < 8 and grid[pr][pc].upper() == 'P':
            if is_attacker(grid[pr][pc], by_white):
                return True

    # 2. Knight attacks (L-shape)
    knight_moves_deltas = [(-2, -1), (-2, 1), (-1, -2), (-1, 2), (1, -2), (1, 2), (2, -1), (2, 1)]
    for dr_k, dc_k in knight_moves_deltas:
        kr, kc = r + dr_k, c + dc_k
        if 0 <= kr < 8 and 0 <= kc < 8 and grid[kr][kc].upper() == 'N':
            if is_attacker(grid[kr][kc], by_white):
                return True

    # 3. King attacks (1 step in any direction)
    # A king cannot move into a square attacked by another king, but for simplicity of check detection, we consider a king's direct vicinity as attacked by an opposing king.
    for dr_k, dc_k in [(-1, -1), (-1, 0), (-1, 1), (0, -1), (0, 1), (1, -1), (1, 0), (1, 1)]:
        kr, kc = r + dr_k, c + dc_k
        if 0 <= kr < 8 and 0 <= kc < 8 and grid[kr][kc].upper() == 'K':
            if is_attacker(grid[kr][kc], by_white):
                return True

    # 4. Rook/Queen attacks (straight lines)
    straight_directions = [(-1, 0), (1, 0), (0, -1), (0, 1)] # Up, Down, Left, Right
    for dr_s, dc_s in straight_directions:
        for step in range(1, 8):
            sr, sc = r + dr_s * step, c + dc_s * step
            if not (0 <= sr < 8 and 0 <= sc < 8): break # Out of bounds
            
            piece_at_sr_sc = grid[sr][sc]
            if piece_at_sr_sc == '.': continue # Path clear
            
            if is_attacker(piece_at_sr_sc, by_white) and \
               (piece_at_sr_sc.upper() == 'R' or piece_at_sr_sc.upper() == 'Q'):
                return True
            else: 
                break # Blocking piece is not an attacker or is own piece, or another piece
    
    # 5. Bishop/Queen attacks (diagonal lines)
    diagonal_directions = [(-1, -1), (-1, 1), (1, -1), (1, 1)] # Up-Left, Up-Right, Down-Left, Down-Right
    for dr_d, dc_d in diagonal_directions:
        for step in range(1, 8):
            sr, sc = r + dr_d * step, c + dc_d * step
            if not (0 <= sr < 8 and 0 <= sc < 8): break # Out of bounds
            
            piece_at_sr_sc = grid[sr][sc]
            if piece_at_sr_sc == '.': continue # Path clear

            if is_attacker(piece_at_sr_sc, by_white) and \
               (piece_at_sr_sc.upper() == 'B' or piece_at_sr_sc.upper() == 'Q'):
                return True
            else: 
                break # Blocking piece is not an attacker or is own piece, or another piece
    return False
\end{lstlisting}

\end{document}